\documentclass[11pt]{google} 
\usepackage[authoryear,round]{natbib}

\usepackage{hyperref}
\usepackage{url}
\usepackage[utf8]{inputenc} 
\usepackage[T1]{fontenc}    
\usepackage{hyperref}       
\usepackage{url}            
\usepackage{booktabs}       
\usepackage{amsfonts}       
\usepackage{nicefrac}       
\usepackage{microtype}      
\usepackage{xcolor}    
\usepackage{subcaption}

\usepackage{latexsym}

\usepackage[T1]{fontenc}

\usepackage[utf8]{inputenc}

\usepackage{microtype}

\usepackage{graphicx}

\usepackage{microtype}
\usepackage{hyperref}
\usepackage{url}
\usepackage{booktabs}
\usepackage{amsmath}
\usepackage{wrapfig}
\usepackage{quoting}
\usepackage{booktabs}
\usepackage{multirow}
\usepackage{makecell}

\usepackage{nicematrix}
\usepackage{tabularray}
\usepackage{longtable}
\usepackage{pifont}
\usepackage{tikz}

\usepackage{amsmath}
\usepackage{amssymb}   

\newcommand{\reminder}[1]{[#1]}
\newcommand{\tianyu}[1]{\reminder{{\bf\small\color{violet} Tianyu: ~#1}}}

\usepackage{tcolorbox}
\usepackage{listings}
\usepackage[dvipsnames]{xcolor}
\usepackage{fancyvrb}
\definecolor{lightblue}{RGB}{46,110,187}
\definecolor{darkred}{RGB}{150,38,31}
\definecolor{darkgreen}{rgb}{0.0, 0.75, 0.0}
\definecolor{blue}{HTML}{3572EF}

\usepackage{lipsum}

\usepackage{arydshln}
\usepackage{subcaption}  
\usepackage{enumitem}

\usepackage{lineno}

\definecolor{darkblue}{rgb}{0, 0, 0.5}
\hypersetup{colorlinks=true, citecolor=darkblue, linkcolor=darkblue, urlcolor=darkblue}

\usepackage{colortbl}
\definecolor{newmodel}{HTML}{F8D7D3}

\renewcommand{\footrule}{\ifnum\value{page}=1 \hrule width 0.4\textwidth height 0.4pt \vskip 3pt\fi}
\renewcommand{\today}{2026-11-15}
\newcommand{\titlebreak}{ }
\title{AI Reviews Are Biased Towards the Appendix}

\makeatletter
\renewcommand\AB@authnote[1]{}
\renewcommand\AB@affilnote[1]{}
\makeatother
\author{Saptarshi Ghosh}
\author{Mohammad Saim}
\author{Tianyu Jiang}
\affil{University of Cincinnati}
\correspondingauthor{ghosh2si@mail.uc.edu, saimmd@mail.uc.edu, tianyu.jiang@uc.edu}

\begin{abstract}
  Large language models (LLMs) are increasingly used by reviewers to highlight weaknesses during the reviewing process, and their outputs often shape the reviews that humans submit. Prior work studies how LLM reviews align with human ones, yet we know little about which parts of a paper these models actually criticize. In this work, we investigate 2,000 papers from ACL, EMNLP, COLM, and ICLR, and show that the latest GPT and Claude models draw about 50\% or more of their review weaknesses from the appendix. This behavior holds across the API, coding-agent, and web access methods for using LLMs to review a paper. In contrast, we find that humans draw only 8.4\% of their weaknesses from the appendix, and these weaknesses primarily revolve around a paper’s presentation and content related to its core idea. Meanwhile, LLMs draw weaknesses from deep in the appendix that concern fine-grained technical details not central to the paper’s contributions. We anticipate that our analysis will contribute to the scholarly discourse on AI reviews. Our code is publicly available at: \url{https://github.com/cincynlp/review_appendix}.
\end{abstract}

\begin{document}

\renewcommand{\titlebreak}{\\}
\maketitle
\renewcommand{\titlebreak}{ }

\begin{figure*}[h]
    \centering
    \includegraphics[width=0.95\linewidth]{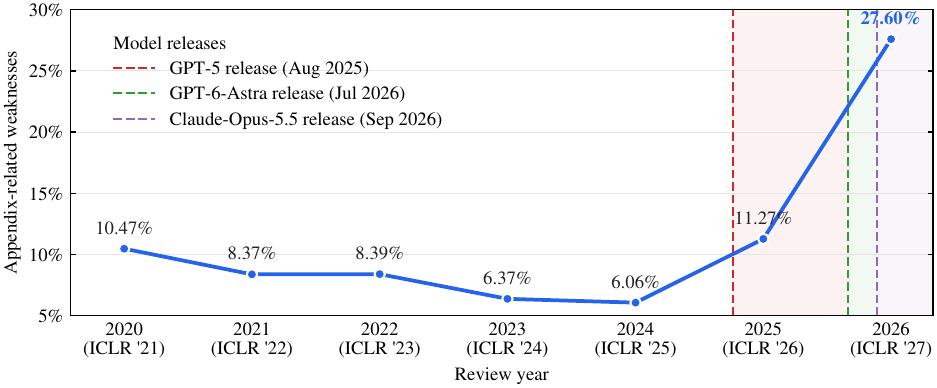}
    \caption{Newer models such as GPT-6-Astra and Claude-Opus-5.5 are biased toward the appendix when writing reviews. They draw more than 50\% of their weaknesses from appendix content. This bias now seems to be showing up in human reviews. For years, human reviewers at ICLR cited only about 6-10\% of their weaknesses in the appendix. This share began to rise after the release of GPT-5, and nearly tripled to 27.60\% in ICLR 2027 reviews, posted shortly after GPT-6-Astra and Claude-Opus-5.5 were released.}
    \label{fig:human_trend}
\end{figure*}

\section{Introduction}

Asking a large language model (LLM) to review a research paper is becoming increasingly common. This is no longer a niche practice but a visible part of peer review. In ICLR 2026, about a fifth of all reviews were flagged as fully AI-generated~\citep{glickenhaus2026pangram4technicalreport}, and many more were likely written with some level of LLM assistance. It is easy to see how this use spreads: reviewers upload a draft to popular AI models like ChatGPT or Claude, ask for a conference-style review, and receive feedback within minutes. The most important part of this feedback is often the list of weaknesses, which reviewers carry into their own reviews.

Recently, we noticed a recurring pattern in high-tier conference review systems. Many criticisms or weaknesses in these AI reviews came from deep in the appendix. Details that were included only as supporting evidence or optional experiments made up a large share of the major weaknesses. This matters beyond AI reviews themselves. Since reviewers often use LLMs to help find weaknesses, what the LLM points out can easily end up in a human review. These weaknesses can also shape a human reviewer’s overall impression of the paper. If an LLM lists multiple weaknesses from the appendix, a reviewer may conclude the paper is weak. Such weaknesses rarely engage with a paper’s core idea, and they do little to help authors improve it. While recent work has studied many aspects of AI review, such as alignment with human reviews~\citep{fichtl2026aiassistedpeerreviewresearch, cameloguerrero2026closelyllmreviewsalign} or benchmarking AI reviews~\citep{li2026llmasareviewerbenchmarkingabilitydivergence, nguyen2026benchmarking}, this bias towards the appendix appears new, emerging mainly with the latest, more powerful models such as GPT-6-Astra and Claude-Opus-5.5.

To this end, we investigate how much of an AI review’s criticism comes from the appendix. To our knowledge, we are the first to study this behavior. We analyze 2,000 papers: 800 from four recent conferences (ACL 2026, EMNLP 2026, COLM 2026, and ICLR 2027) and 1,200 from ICLR 2021 through 2026. We use four state-of-the-art models that were widely used for AI review at the time of this analysis: GPT-6-Astra, GPT-6.1-Sol, Claude-Opus-5.5, and Claude-Sonnet-5.5. Across the four recent conferences, our results show that almost all models draw more than half of their weaknesses from appendix-related material. Compared with earlier models like GPT-5, where less than 30\% of weaknesses relate to the appendix, GPT-6-Astra and GPT-6.1-Sol show around 60-70\%, suggesting this behavior is new. We also study human-written reviews from ICLR 2021 to 2027. We find that humans focus mainly on the main body and draw less than 10\% of their weaknesses from the appendix. This share rises sharply to 27.60\% in human reviews written after the release of newer models, as shown in Figure~\ref{fig:human_trend}. While human reviewers distinguish between the main body and the appendix and weigh them differently in their overall review, LLMs seem more biased toward the supplementary material and treat the paper as a single unit. As more people use these models to review papers, we believe the community should take note of this behavior. In summary, our contributions are:

\begin{itemize}[noitemsep, topsep=2pt, leftmargin=*]
    \item \textbf{Appendix-Heavy Criticism.} We show that the latest GPT and Claude models draw more than 50\% of their review weaknesses from the appendix. This behavior holds across models, venues, and access methods, including the API, CLI agent, and web interface.

    \item \textbf{Comparison with Human Reviews.} In contrast to LLMs, humans drew only 8.4\% of their weaknesses from the appendix in previous years. However, with the arrival of newer models, appendix-related weaknesses in recent human reviews have risen to 27.60\%.

    \item \textbf{Analysis of Appendix Criticism.} Unlike humans, LLMs rarely engage with a paper's core contribution when criticizing the appendix. Instead, they focus on small statistical details and ask for ever more experiments or statistical validation on existing results.
\end{itemize}

\begin{figure*}[t]
    \centering
    \includegraphics[width=0.99\linewidth]{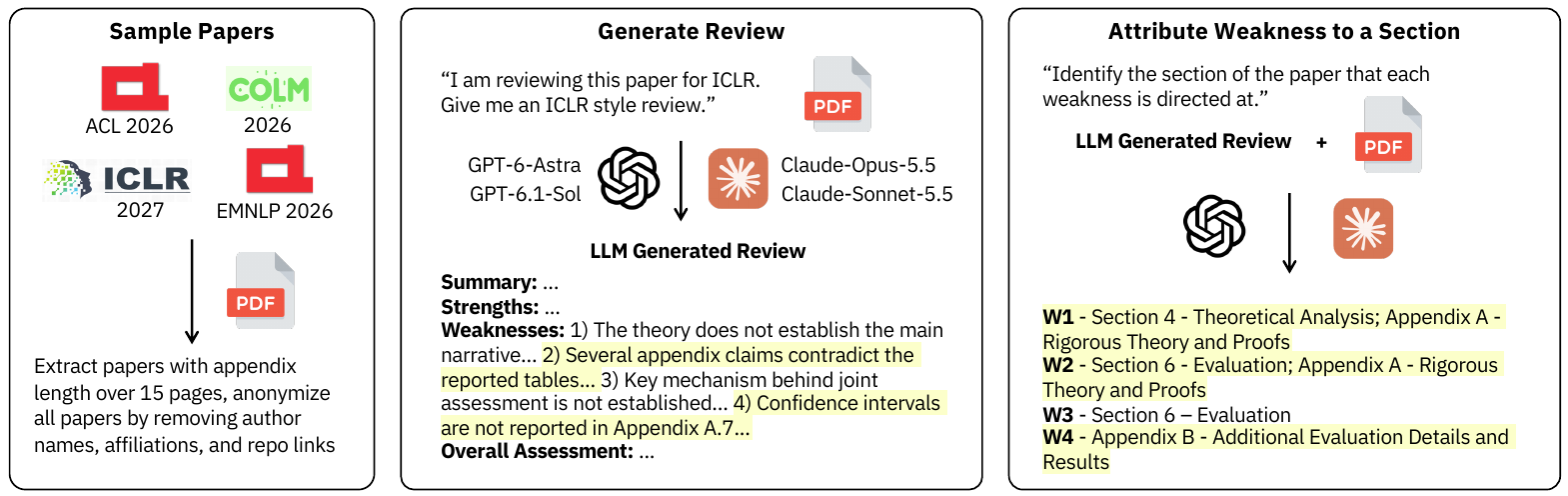}
    \caption{Pipeline of our method: i) Extract and sample papers from four conferences with a minimum appendix length of 15 pages, ii) Generate an AI review, and iii) map each weakness back to the section it belongs to.}
    \label{fig:method}
\end{figure*}

\section{Related Work}

Much of the literature on AI review is recent. Prior work showed that LLMs can provide useful feedback on research papers and studied how reliable they are as reviewers~\citep{liang2023largelanguagemodelsprovide, zhou-etal-2024-llm, du-etal-2024-llms}. Moreover, ample work compares LLM reviews with human reviews and finds that they agree on broad outcomes but differ in what they criticize~\citep{cameloguerrero2026closelyllmreviewsalign, żurawicki2026praibpeerreviewai, chen-etal-2026-peercheck}. Several benchmarks evaluate LLMs and agentic systems as reviewers~\citep{li2026llmasareviewerbenchmarkingabilitydivergence, nguyen2026benchmarking, teo2026beyond}, and a recent survey summarizes this growing area~\citep{nguyen2026llmbasedscientificpeerreview}. At the same time, LLM use in real peer review is rising, and studies estimate how many reviews are written or edited with LLMs~\citep{kumar-etal-2025-mixrevdetect, liang2026monitoringaimodifiedcontentscale,
shen2026detectingaigeneratedcontentacademic}. Studies have also examined whether reviewer policies on AI use can be enforced~\citep{saha2026policies, fichtl2026aiassistedpeerreviewresearch} and measured how LLM use affects review outcomes~\citep{kim2026useeffectsllmspeer,sharma2026llmsfavorllmsquantifying}. Others have put forward the notion that AI should support human reviewers rather than replace
them~\citep{baumann2026stop, yun2026positionllmassistedpeerreview}.

Studies have also highlighted the existing issues of AI review. LLM reviewers can be manipulated through hidden prompts in manuscripts~\citep{lin2025hidden, li2026llmasareviewerbenchmarkingabilitydivergence} or simply by rewording and restyling a paper~\citep{kaneko2026paraphrasingadversarialattackllmasareviewer,
yang2026no}. ~\citet{vasu2026stylesubstancelinguisticbias} showed that they favor confident writing over substance. Together, these studies examine many aspects of AI review, from quality and alignment to bias and manipulation. However, AI reviewers’ tendency to draw criticism from the appendix is a recent trend that has emerged with the latest models and has not yet been studied. To
our knowledge, this work is the first to examine it. As the latest models can now read every page of a paper and are increasingly used by humans to assist with reviewing, this question has become urgent.

\section{Methodology}

In this section, we describe how we extract research papers and our methodology.

\noindent \textbf{Models.} For evaluation, we select four state-of-the-art LLMs widely used for reviewing papers from two model families: GPT-6-Astra~\citep{openai2026gpt6astra} and GPT-6.1-Sol~\citep{openai2026gpt61sol}; Claude-Opus-5.5~\citep{claude_opus} and Claude-Sonnet-5.5~\citep{claude_sonnet}. We use Codex and Claude Code to access their respective models and test whether their behavior aligns with other methods, such as the API and the web interface.

\subsection{Collecting Research Papers}

\noindent \textbf{Conference selection.}
We collect papers from four major machine learning and NLP conferences: ACL 2026, EMNLP 2026, COLM 2026, and ICLR 2027. We chose these venues because their papers and reviews follow a similar structure.
Each paper has a page-limited main body, with extra material placed in appendices after the references. Appendices usually do not have a page limit. The review system typically asks reviewers to include a summary, strengths, weaknesses, numerical scores or ratings based on key parameters, and an overall assessment score. This shared format lets us compare results across venues on an equal footing. We use only recent editions (2026 onward) to avoid cases where LLMs have seen these papers during training.

\noindent \textbf{Sampling \& Anonymizing Papers.}
From each conference, we sample 200 papers with appendices of at least 15 pages. The final sample hence consists of 800 unique papers. The ICLR 2027 papers are anonymous submissions, while the ACL, EMNLP, and COLM papers are camera-ready versions of accepted papers. To match a double-blind review setting, we anonymize the accepted papers by removing author names, affiliations, and email addresses from the PDFs. We also redact all links, such as code repositories or project pages, since these often reveal author and institution identities (left panel in Figure~\ref{fig:method}). Finally, in our main experiments, we disable web search and other browsing tools for all LLM reviewers so the models cannot look up the paper or its authors online and must rely only on the provided PDF.

\subsection{Generating Reviews}

We use a two-step pipeline for this task. We first generate a review for each paper, then identify which part of the paper each weakness comes from.

\noindent \textbf{Review generation.}
We give each of the anonymized papers to the LLM with a simple prompt:
``\textit{I am reviewing this paper for [conference\_name]. Give me a [conference\_name] style review.}'' We keep the prompt short and simple on purpose.
This mirrors how a reviewer might plausibly use an LLM in practice, and avoids confounding the generated review with external prompt artifacts. Because all four conferences use a similar review format, the model can produce a structured review (summary, strengths, weaknesses, and scores) from the conference name alone, without us spelling out the format (Figure~\ref{fig:method} middle panel).

\noindent \textbf{Section attribution.}
Once we have the generated reviews, we want to know which parts of the paper the model draws its weaknesses from. To localize that, we give the paper and its generated review back to the same model and ask it to name the section of the paper each weakness comes from (right panel in Figure~\ref{fig:method}). Some points span more than one section, such as ``\textit{the claims made in the introduction are not supported by the experiments...},'' so we allow the model to name up to two sections. Some points are general and do not refer to any specific section, such as ``\textit{the paper is well written}.'' For these, we instruct the model to answer ``not specified.''  

We split the pipeline into two steps on purpose. An alternative would be to ask the model to name the section for each weakness while generating the review itself. However, this would add extra instructions to the prompt and make it less natural. Our design keeps the prompt close to real-world usage, so prompt artifacts do not affect the generated review.

\section{Experimental Results}

We assign each weakness to either the main body or the appendix. A weakness is linked to the appendix if any part of its supporting evidence comes from the appendix, whether from the appendix alone (W4 in Figure~\ref{fig:method}) or from the appendix and the main body together (W2 in Figure~\ref{fig:method}). We then compute the percentage of weaknesses in the main body and the appendix.

\subsection{LLMs Draw Heavily on the Appendix}

\begin{figure*}[t]
    \centering
    \includegraphics[width=0.98\linewidth]{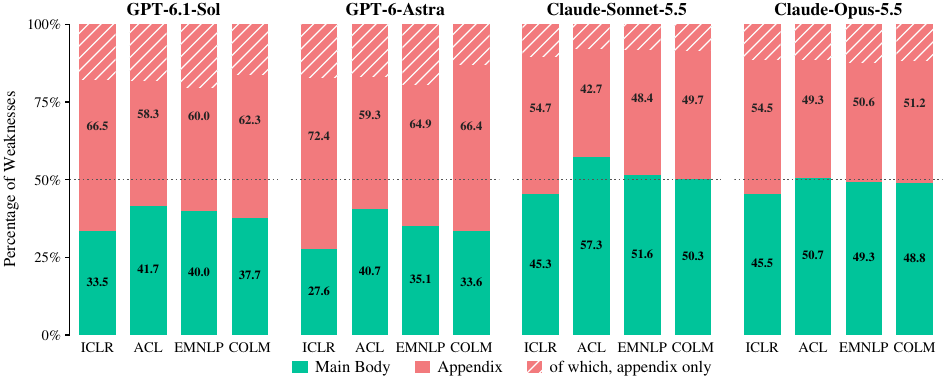}
    \caption{Percentage of weaknesses in LLM-generated reviews drawn from the main body, appendix, and both combined, from recent conferences (ACL 2026, EMNLP 2026, COLM 2026, and ICLR 2027). Across almost all, more than half of the weaknesses are drawn from appendix-related content.}
    \label{fig:main_weakesses}
\end{figure*}

Figure~\ref{fig:main_weakesses} shows where the weaknesses in LLM-generated reviews come from. The pattern is consistent across all models and all four conferences: about half or more of the weaknesses draw on the appendix. GPT draws 17.33\% of its weaknesses from the appendix alone (white shade), with no link to the main body, compared to 10.16\% for Claude. The results indicate that LLM reviewers do not distinguish between the weight of the main-body content and the appendix. A large share of their criticism comes from content that authors often treat as secondary. For these reviewers, the appendix is not optional reading. It directly shapes the feedback a paper receives.

\subsection{Comparing Model Variants}

We repeat our analysis with GPT-5, an earlier model from the same provider, and compare it against the newer GPT-6-Astra and GPT-6.1-Sol. This lets us directly compare across model generations to observe changes in preference or model bias when reviewing the paper. We run GPT-5 through the same two-step pipeline and, for each model, compute the share of weaknesses that involve the appendix.

\begin{figure}[t]
    \centering
    \begin{subfigure}[t]{0.49\linewidth}
        \centering
        \includegraphics[width=\linewidth]{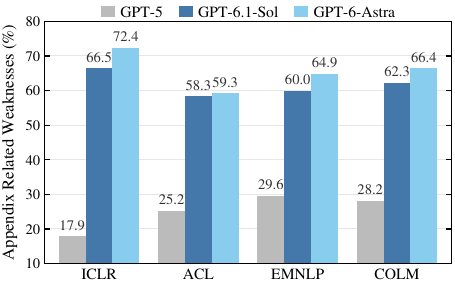}
        \caption{Percentage of appendix-related weaknesses. It rises in the latest variants, GPT-6.1-Sol and GPT-6-Astra, while GPT-5 had less.}
        \label{fig:earlier_weakness}
    \end{subfigure}
    \hfill
    \begin{subfigure}[t]{0.49\linewidth}
        \centering
        \includegraphics[width=\linewidth]{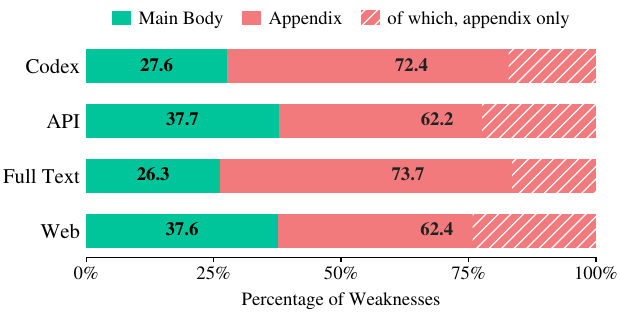}
        \caption{Where weaknesses stem from, compared across access methods of GPT-6-Astra.}
        \label{fig:methods}
    \end{subfigure}
    \caption{Overview of appendix-related weaknesses: (a) trend across model variants, (b) breakdown by access method.}
    \label{fig:weakness_overview}
\end{figure}

Figure~\ref{fig:earlier_weakness} shows the results. The difference between the generations is large. For GPT-5, only a small share of weaknesses relate to the appendix. Both newer models, GPT-6.1-Sol and GPT-6-Astra, rely on it much more heavily, and this holds for every conference. This suggests that the \textbf{bias towards the appendix is mainly a property of newer models}. Even though the new models are an improvement, their bias toward the appendix has grown. Being able to read a paper in detail has not taught them which parts matter most.

\subsection{Comparing Access Methods}

There exist different approaches for using OpenAI or Anthropic products to review a paper. In daily use, most people upload the PDF to the web (chat interface) or use coding agents such as Codex. On the other hand, researchers may also call the model through an API, which is easier to control and repeat at scale.  While these methods use the same underlying model, they differ in how they read the paper.

\noindent \textbf{API.} With the API, the whole PDF is sent to the model in a single request, so it sees the full paper at once, including figures, tables, and layout. Another common way researchers use the API is to extract the text from the PDF themselves and feed it to the model. This is not ideal for paper review, since the model sees the paper without its figures. Nonetheless, we include it because it removes every layer between the model and the paper (no PDF parsing by the API). The model receives the full text at once, making this setting a clean control for the model's own behavior.

\noindent \textbf{Codex.} Codex works differently as it does not receive the paper directly. It runs shell commands to read the PDF, usually starting with a text extraction tool such as pdftotext, and reads the command’s output. However, Codex limits how much command output it passes back to the model. By default, this limit is about 10k tokens per tool call. When the output is longer, Codex keeps the beginning and the end, drops the middle, and inserts a short note saying how many tokens were removed. This cut is made by Codex harness, not by the model. To see the rest, the model has to ask for it with follow-up commands, such as searching for section headings or extracting specific page ranges. What it ends up reading is therefore what it chooses to look up, and it may skip some sections. 

\noindent \textbf{Web.} The web framework for accessing the model may also vary as it could include varying system prompts, memory, and safety layers to optimize for user dialogue. Therefore, the model response can vary based on the added processing that users cannot configure. The model's method for reading and analyzing an uploaded file is not public. We provide one PDF at a time and turn off memory.

In short, the methods differ in how the model reads the paper. This could change how much attention the appendix receives; therefore, we test all four access methods. We run the same review pipeline on 100 ICLR 2027 submissions using GPT-6-Astra with four methods: Codex, the full extracted text as API input, API (direct PDF input), and the web interface.

\noindent \textbf{Results.} Figure~\ref{fig:methods} shows the results. Appendix dependency is consistent across methods, with all methods showing more than 50\% of weaknesses from appendix-related material. Although each method views the paper differently, bias toward the appendix is consistent. All methods cite weaknesses from the appendix, suggesting this behavior reflects the underlying model.

\section{Human vs LLM Reviews: Appendix Dependency Study}

So far, we have seen that LLM-generated reviews draw heavily on the appendix for weaknesses. Since human reviewers can also draw weaknesses from the appendix, this raises the question: how do humans and LLMs differ in the way they draw weaknesses from the appendix? In this section, we study this question by comparing human-written reviews with LLM reviews.

\noindent \textbf{Extracting Human Reviews.} To compare LLM with human review, we collect human-written reviews for ICLR 2021 to 2024 from OpenReview. Each ICLR edition is reviewed in the previous calendar year. For example, ICLR 2021 papers were reviewed in 2020. For each year, we sample 200 papers using the same rule as before: the paper must have an appendix of at least 10 pages. Long appendices were less common in earlier editions of the conference, so we keep the limit at 10 pages. We choose the years 2021-2024 on purpose. Before 2021, the conference received fewer submissions, and appendices were usually short. From 2025 onward, reviewers may well have used LLMs to help write their reviews, which would blur the line between human and LLM reviews and confound our results. The intervening years give us a clean window. Each paper typically has 3-4 reviews, and we randomly select one review per paper.

To identify the source of each weakness, we reuse the second step of our pipeline (right panel in Figure~\ref{fig:method}) to attribute each weakness to a specific section. We give GPT-6.1-Sol the PDF and the corresponding human-written review and ask it to name the paper section each weakness refers to. We then compare the results with 200 ICLR 2027 papers whose reviews were generated by LLMs. We do not directly compare human and LLM-generated reviews for the ICLR 2021-2024 papers themselves because these papers and their reviews have been public on OpenReview for years and have likely been part of the models’ training data. ICLR 2027 submissions avoid this problem (as submissions to this conference have been close to the time of model release).

\subsection{Humans Prioritize the Main Body, LLMs Do Not}

\begin{wraptable}{r}{0.42\textwidth}
    \vspace{-12pt}
    \centering
    \caption{Appendix-related weaknesses (\%) in human reviews (ICLR 2021--24) and LLM reviews (ICLR 2027).}
    \label{tab:human_overview}
    \small
    \begin{tabular}{lcc}
    \toprule
     & \shortstack{Appendix-\\related} & \shortstack{Appendix\\only} \\
    \midrule
    Humans & 8.40 {\scriptsize $\pm$1.44} & 4.36 {\scriptsize $\pm$1.03} \\
    LLMs   & \textbf{62.00} {\scriptsize $\pm$7.70} & \textbf{14.19} {\scriptsize $\pm$3.32} \\
    \bottomrule
    \end{tabular}
    \vspace{-8pt}
\end{wraptable}
Table~\ref{tab:human_overview} shows the percentage of appendix-related weaknesses in human reviews (ICLR 2021-2024) and LLM reviews (ICLR 2027). For LLM reviews, we report the mean over all four LLMs in Figure~\ref{fig:main_weakesses}. The overall picture is clear: human reviewers treat the main body and the supplementary material very differently from LLMs. LLMs attribute more than 50\% of their weaknesses to appendix-related material, while humans attribute less than 10\%. The difference also shows at the individual-review level. Of the 200 papers sampled each year from 2021 to 2024, only 159 in total received a human review with any appendix-related weakness. In contrast, all 200 LLM-generated reviews for ICLR 2027 included at least one appendix criticism.

\noindent \textbf{A Growing Trend in Human Reviews.} If reviewers are using LLMs to help write their reviews, we would expect human reviews to slowly adopt the response of these models in their posted reviews. To test this, we track how often human-written weaknesses involve the appendix across 1,400 papers and reviews from ICLR 2021-27 (200 papers from each year).  

Table~\ref{tab:human_trend} shows the result (also Figure~\ref{fig:human_trend}). Appendix-related weaknesses remained largely the same from 2020 to 2024. LLMs were already being used in peer review by 2023~\citep{liang2026monitoringaimodifiedcontentscale}, but human reviews did not shift toward the appendix. This changed in the past couple of years (2025 and 2026). The share of appendix-related weaknesses rises sharply in ICLR 2026 reviews, the first review cycle after the release of GPT-5. This is also the cycle in which about a fifth of reviews were flagged as fully AI-generated \citep{glickenhaus2026pangram4technicalreport}. The numbers rise drastically in 2026 to 27.60\%, when the review period followed the release of GPT-6-Astra and Claude-Opus-5.5. The timing points to a simple explanation: reviewers did not suddenly start reading appendices more closely, but the tools they relied on did.

\begin{table}[t]
    \centering
    \caption{Percentage of appendix-related weaknesses in human-written
    reviews for each ICLR edition from 2021 to 2027. Appendix-related
    weaknesses increase sharply after the release of the current models.
    $^{*}$Reviews written after the release of the evaluated models.}
    \label{tab:human_trend}
    \resizebox{0.9\linewidth}{!}{
    \begin{tabular}{lccccccc}
    \toprule
    Review year & 2020 & 2021 & 2022 & 2023 & 2024 & 2025 & 2026 \\
    \midrule
    Appendix-related weaknesses (\%) & 10.47 & 8.37 & 8.39 & 6.37 & 6.06 & 11.27 & 27.60$^{*}$ \\
    \bottomrule
    \end{tabular}}
\end{table}

\begin{wrapfigure}{r}{0.50\textwidth}
    \vspace{-10pt}
    \centering
    \includegraphics[width=0.49\textwidth]{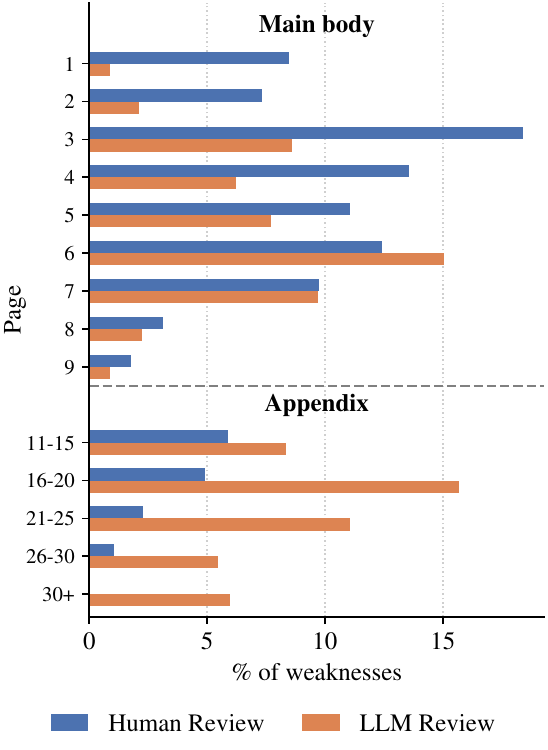}
    \caption{Percentage of weaknesses from each page, comparing human and LLM reviews. Human reviews are from ICLR 2021--24, LLM reviews are from ICLR 2027.}
    \label{fig:page}
    \vspace{-10pt}
\end{wrapfigure}

\noindent \textbf{Where Do Weaknesses Come From?} Table~\ref{tab:human_overview} gives the big picture. To look more closely at how human and LLM reviews differ, we ask which pages of a paper each weakness comes from. The second stage of our pipeline identifies the section that each weakness refers to (e.g., \textit{W4 - Appendix D - Experiment Details}). We map this section back to its page number in the PDF, which gives a finer view of where reviewers focus. We then compute the percentage of weaknesses from each page across both the main body and the appendix. For human reviews, we use only the reviews that contain at least one appendix-related weakness. For LLM reviews, we use all reviews.

Figure~\ref{fig:page} shows the percentage of weaknesses from each page for human and LLM reviews. The results are intuitive. Human reviewers cite more weaknesses from the first five pages. This is where authors usually present their goals and motivation, core idea, and method. Humans focus their criticism on the core ideas of the paper. LLMs, by contrast, cite more weaknesses from pages 6 and 7, which mostly contain results and analysis, and, importantly, from throughout the appendix.  

Two patterns stand out. First, LLMs almost never criticize the opening pages, where a paper frames its problem and argues why it matters. Humans often push back on exactly this framing, while LLMs seem to accept it and move on to the numbers. Second, the two groups read the appendix very differently. Human criticisms fade quickly as the appendix goes on, and almost disappear in its last pages. LLM weaknesses do not fade at all. In fact, LLM weaknesses peak well into the appendix rather than at its start, and they continue even beyond page 30, where human reviewers raise almost no concerns. For an LLM, a detail on page 25 is as fair a target as the main results table.

\subsection{How Do Human and LLM Criticisms of the Appendix Differ?}

Having established that humans prioritize the main body while LLMs target the appendix as much as the main body, we next ask how humans and LLMs criticize a paper’s appendix. This is important because the amount of criticism alone does not tell the whole story. A weakness drawn from the appendix can be useful, such as catching a real error in an experiment, or it can be minor, such as pointing out a missing detail that does not affect the paper’s main claims. Understanding this difference is important in establishing how LLMs and humans differ in reading and criticizing the appendix.


\begin{table}[t]
    \centering
    \caption{Categories of appendix-related weaknesses in human reviews (ICLR 2021-2024) and LLM reviews (ICLR 2027).}
    \resizebox{\linewidth}{!}{
    \begin{tabular}{lcccccccc}
    \toprule
     & Methodology & Results & Analysis & Presentation & Prompt & Discussion & Tables \& Fig. & Core idea \\
    \midrule
    Humans & 35.7 & 6.5 & 13.5 & 32.2 & 1.3 & 0.3 & 7.8 & 1.7\\
    LLMs  & 54.0 & 13.6 & 25.2 & 2.1 & 0.3 & 0.3 & 3.0 & 1.4 \\
    \bottomrule
    \end{tabular}}    
    \label{tab:human_llm_taxonomy}
\end{table}

\noindent \textbf{Weakness Taxonomy.} To understand what humans and LLMs criticize in the appendix, we sort each appendix-related weakness into one of eight categories: \textit{Core Idea}, \textit{Methodology}, \textit{Results}, \textit{Analysis}, \textit{Prompts}, \textit{Discussion}, \textit{Tables and Figures}, and \textit{Presentation}. We collect weaknesses that cite the appendix, either alone or with the main body, and ask GPT-6.1-Sol to assign each to a single category based on what it criticizes. If no category fits, the model can answer \textit{None}. The model is instructed to rely only on what the review says and what the paper contains, and not to rewrite or reassess the review. We then compute the percentage of weaknesses in each category.

Table~\ref{tab:human_llm_taxonomy} shows what humans and LLMs criticize when they draw on the appendix. Both focus on Methodology, but LLMs do so far more, and they also raise more concerns about Results and Analysis. A large share of human criticisms goes to Presentation, which concerns the paper’s organization and readability, a category LLMs almost never use. Overall, LLM criticism of the appendix is around technical details, while human criticism also covers how the paper is written and organized.

\noindent \textbf{Qualitative Analysis.} The categories tell us what reviewers criticize, but not how. To get a closer look, we read through appendix-related
weaknesses from the main categories and compare how humans and LLMs express them. Table~\ref{tab:qualitative_analysis} shows some examples. On methodological details, human reviewers tend to dig deep into a single issue \textit{that matters to the core of the paper}. LLMs treat the appendix as technical material to audit, highlighting details unrelated to the paper’s central idea. A major factor is human expectations for presentation. Humans expect the main body to stand on its own, so they object when essential sections are pushed into the appendix. LLMs almost never raise this concern. This highlights a key difference: humans recognize that the main body carries more weight than the appendix, while LLMs do not. Another important observation is how LLMs criticize the results and analysis. They cross-check the main text against appendix details to find small mismatches, question whether the results truly support the claims, and suggest confounds or alternative explanations. They also repeatedly ask for more experiments and statistical checks, such as additional seeds, significance tests, confidence intervals, etc. These points are often valid, but they do not engage with the paper’s core contribution or main idea.

\definecolor{rowblue}{RGB}{134, 211, 246}
\definecolor{rowyellow}{RGB}{240, 228, 119}

\begingroup
\renewcommand\baselinestretch{0.99}
\begin{table*}[t]
    \NiceMatrixOptions {
    custom-line = {
       command = dashedmidrule ,
       tikz = { dashed } ,
       total-width = \pgflinewidth + \aboverulesep + \belowrulesep ,
     } }
    \caption{Qualitative analysis of human and LLM weaknesses attributed to the appendix showing how different the weaknesses are from the primary categories in Table~\ref{tab:human_llm_taxonomy}.}
    \centering
    \small
    \resizebox{\textwidth}{!}{
    \begin{NiceTabular}{|p{0.45\linewidth}|p{0.55\linewidth}|}
    \CodeBefore
      \rowcolor{rowblue!10}{3,5,7}
    \Body
    \toprule
    \multicolumn{1}{c}{\bf  Human Review} & \multicolumn{1}{c}{\bf LLM Review} \\
    \midrule
    \textbf{Methodology:} Investigates a specific methodological issue in depth & \textbf{Methodology:} Cross-checks main-body claims against appendix details, highlighting implementation-level mismatches. \\ 
    \textbf{Example:} \textit{Since Algo.2 is using a non-trivial averaging for the normalization term, the convergence of the stochastic version is unclear...} & \textbf{Example:} \textit{Equation (9) weights the routing KL by T, whereas Appendices B.1 and B.3 report a KL weight of 1.0 without explaining how this corresponds to the derived objective}\\ 
    \midrule
    \textbf{Presentation:} Placement of essential content, essential intuition buried in the appendix & \textbf{Results \& Analysis:} Scrutinizes whether results establish the claims, highlighting confounds and alternative explanations for improvements. \\ 
    \textbf{Example:} \textit{...this paper especially abuses the appendix often sending important parts of the discussion and intuition for the results into appendix G.}
    & 
    \textbf{Example:} \textit{...the observed effects might be a result of pragmatic intensity...the paper does not isolate the effect of pragmatic intensity from figurative language which might confound the results.}
    \\ 
    \midrule
    \textbf{Presentation:}  Criticizes the main text for depending on the appendix for key explanations and procedures  & \textbf{Results \& Analysis:} Demands additional experiments and analyses, such as ablations, controls, repeated seeds, variance estimates, significance tests, and confidence intervals.  \\
    \textbf{Example:} \textit{...the main text of the paper intensely refer to contents in the appendix, including the actual testing procedure pseudocode...the current text without these appendix is not self-contained.} & 
    \textbf{Example:} \textit{Appendix A does not provide any statistical significance test or confidence interval... repeated runs and confidence intervals are also needed to interpret small ablation differences... multiple seeds and paired confidence intervals is required...}\\
    \bottomrule
    \end{NiceTabular}
    }
    \label{tab:qualitative_analysis}
\end{table*}
\endgroup

\section{Discussion}

\noindent \textbf{Our Interpretation.} We offer our interpretation of why this behavior occurs. Reading everything is not the same as weighing everything. Human reviewers tend to follow an unwritten rule: the main body is the paper, and the appendix exists to support it. Authors therefore focus primarily on polishing the main content, and usually refer to additional experiments or more technical details in the appendix. LLMs, however, seem to weigh every line equally, which leaves them more likely to find weaknesses in the appendix. The appendix is usually longer, less polished, and full of details such as hyperparameters, extra tables,
and edge cases, which makes it easy ground for finding flaws. For a model trained to be thorough and specific, the appendix becomes a gold mine of
weaknesses.

\noindent \textbf{Is Appendix Bias Good or Bad?} We view appendix bias as having both benefits and drawbacks. On the positive side, authors are responsible for every page they submit. An error in a table in Appendix~G is still an error, and a reviewer who identifies it improves the paper. In this sense, LLM reviewers raise the standard, since they leave no part of the paper unexamined. However, effective reviewing also requires judgment about what matters most. When most of a review’s weaknesses come from appendix content unrelated to the main contribution, the feedback stays on the surface. It does not engage with the core of the paper and does little to help authors improve it. This question is not only academic for us. Our own recent submissions have received reviews where most main concerns came from deep in the appendix, and that experience motivated this work. We hope our findings bring this behavior to light and encourage reviewers to use LLMs more carefully, as a tool for checking the details, not as the judge of what matters most in a paper.

\section{Conclusion}

We show that the latest LLMs derive the majority of their criticism from the appendix. This pattern is consistent across models, conferences, and access methods, and we did not observe this behavior in earlier models. Human reviewers weigh the main body and the appendix differently, but LLMs treat the whole paper as one unit. This matters because reviewers use these models to find weaknesses, and authors use them to judge whether their paper is ready. When a model fills a review with issues from supplementary materials, it can make a paper proposing a strong idea look unpolished and pull attention away from its main contribution. We hope this work helps reviewers use LLMs more carefully, helps authors read AI feedback with the right weight in mind, and encourages model developers to build reviewers that know what matters most in a paper.

\section*{Limitations}

\noindent \textbf{LLM-based attribution.} We use an LLM to trace each weakness back to the section of the paper it comes from, and to sort appendix-related weaknesses into categories. While this lets us analyze thousands of reviews, the model can make mistakes, for example, when a weakness is vague or spans several sections. We mitigate this by instructing the model not to make assumptions and to answer \textit{None} when it is uncertain. We also manually reviewed a sample of the outputs to ensure the attributions are credible (a point we suspect an LLM reviewer would flag for lacking statistical tests). Nonetheless, this remains a limitation. 

\noindent \textbf{Comparing humans and LLMs.} Our human reviews and LLM reviews come from different papers and different years. We made this choice on purpose, since older papers and their reviews are likely part of the models' training data. Still, writing styles and appendix norms change over time, and these differences may affect the comparison. We also note that the rise of appendix-related criticism in human reviews is a temporal correlation, not proof that LLMs caused it.

\noindent \textbf{What we do not measure.} We study where criticism comes from and what it is about, but not whether each weakness is correct or useful. Some appendix-related weaknesses point to real problems rather than surface-level criticisms. We think a natural follow-up work would be to study the validity of weaknesses attributed to the appendix, which we leave for future work. 

Additionally, our focus in this work is on the weaknesses pointed out by the LLM. We do not explore the numerical scores assigned to a paper. Future work can explore how appendix-related weaknesses affect the quantitative scores assigned to a paper.

\section*{Ethics Statement}

\noindent \textbf{Data.} All papers and human reviews in this study are publicly available. We collected papers from ACL, EMNLP, and COLM public proceedings, and ICLR submissions and reviews from OpenReview. ICLR reviews are published anonymously, and we analyze them only in aggregate. We do not try to identify reviewers, and we do not judge whether any individual review was written with LLM help. Our findings on human reviews describe trends across years, not the behavior of specific people.

\noindent \textbf{Generated reviews.} We generate reviews only to study model behavior. We did not use them in any real review process. For accepted papers, we removed author names, affiliations, and links before giving them to the models, to match the double-blind setting of peer review.

\noindent \textbf{Intended use.} Our goal is to help the community understand how AI reviewers read a paper, so that these tools can be used more carefully. We do not suggest that authors should shorten their appendices or leave out supporting evidence. Detailed appendices support reproducibility and remain valuable. Instead, we hope our findings encourage reviewers to weigh appendix-based criticism
appropriately.

\bibliography{ref}
\bibliographystyle{abbrvnat}

\end{document}